\documentclass[10pt,twocolumn,letterpaper]{article}
\usepackage[pagenumbers]{cvpr}

\usepackage[dvipsnames]{xcolor}
\usepackage{algorithm, algorithmic}

\newcommand{\comm}{{\hfill \(\triangleright\)}}
\newcommand{\gray}{\colorbox{lightgray}}
\DeclareMathOperator*{\argmin}{arg\,min}

\newcommand{\x}{{\mathbf{x}}}
\newcommand{\y}{{\mathbf{y}}}

\newcommand{\xzero}{{\mathbf{x}_{0}}}
\newcommand{\xt}{{\mathbf{x}_{t}}}
\newcommand{\xT}{{\mathbf{x}_{T}}}
\newcommand{\xzt}{{\hat{\mathbf{x}}_{0}(\mathbf{x}_{t})}}
\newcommand{\xztt}{{\hat{\mathbf{x}}_{0}^{'}(\mathbf{x}_{t})}}

\newcommand{\xreg}{{\mathbf{x}_{reg}}}

\newcommand{\dm}{{\mathbf{\epsilon}_{\mathbf{\phi}}}}
\newcommand{\eps}{{\mathbf{\epsilon}}}
\newcommand{\gauss}{{\mathcal{N}(\mathbf{0}, \mathbf{I})}}
\newcommand{\dmh}{{\mathbf{\epsilon}^{'}_{\mathbf{\phi}}}}

\newcommand{\thb}{{\mathbf{\theta}}}

\newcommand{\bb}{{\mathbf{\beta}}}
\newcommand{\bc}{{\mathbf{c}}}

\definecolor{cvprblue}{rgb}{0.21,0.49,0.74}
\usepackage[pagebackref,breaklinks,colorlinks,citecolor=cvprblue]{hyperref}

\title{ Score-Guided Diffusion for 3D Human Recovery }

\author{Anastasis Stathopoulos\\
Rutgers University
\and
Ligong Han\\
Rutgers University
\and
Dimitris Metaxas\\
Rutgers University
}

\begin{document}

\twocolumn[{
	\renewcommand\twocolumn[1][]{#1}
	\maketitle
	\begin{center}
            \vspace{-0.15in}
		\newcommand{\teaserwidth}{\textwidth}
		\centerline{
			\includegraphics[width=\teaserwidth,clip]{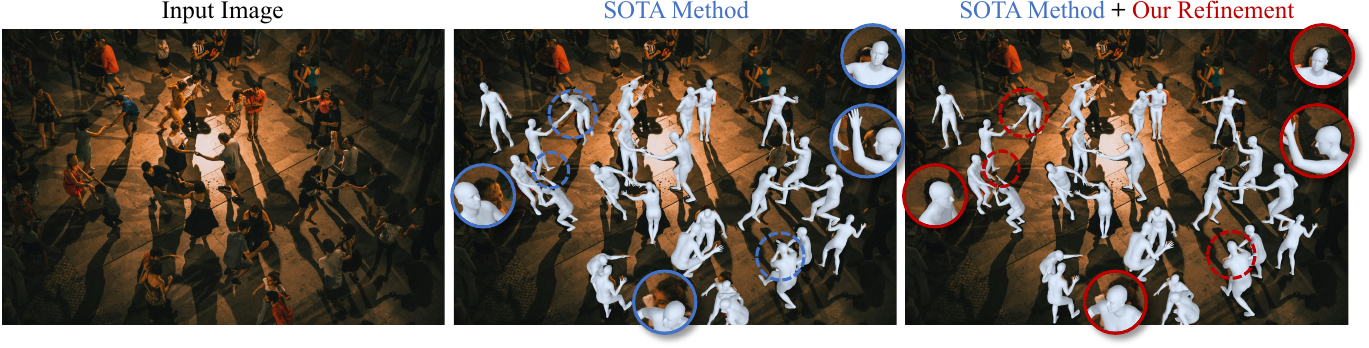}
		}
            \vspace{-0.06in}
		\captionof{figure}{Although achieving remarkable 3D human reconstructions, a recent state-of-the-art monocular regression approach~\cite{goel2023humans} may encounter challenges in aligning the human body model to the image (middle image). To address this, we propose an iterative refinement approach that utilizes image observations (\eg, 2D keypoint detections) and achieves better image-model alignment (right image).
        }
		\label{fig:teaser}
	\end{center}
}]
\begin{abstract}
We present Score-Guided Human Mesh Recovery (ScoreHMR), an approach for solving inverse problems for 3D human pose and shape reconstruction. These inverse problems involve fitting a human body model to image observations, traditionally solved through optimization techniques. ScoreHMR mimics model fitting approaches, but alignment with the image observation is achieved through score guidance in the latent space of a diffusion model. The diffusion model is trained to capture the conditional distribution of the human model parameters given an input image. By guiding its denoising process with a task-specific score, ScoreHMR effectively solves inverse problems for various applications without the need for retraining the task-agnostic diffusion model. We evaluate our approach on three settings/applications. These are: (i) single-frame model fitting; (ii) reconstruction from multiple uncalibrated views; (iii) reconstructing humans in video sequences. ScoreHMR consistently outperforms all optimization baselines on popular benchmarks across all settings. We make our code and models available at the \href{https://statho.github.io/ScoreHMR}{project page}.
\end{abstract}    
\section{Introduction}
\label{sec:intro}

\phantom{11}
Approaches for recovering the 3D human pose and shape from 2D evidence (\eg, image, 2D keypoints) typically predict the parameters of a human body model, such as SMPL~\cite{loper2015smpl}, and solve the problem with regression~\cite{kanazawa2018end,georgakis2020hierarchical,zhang2021pymaf,kocabas2021pare,goel2023humans} or optimization~\cite{bogo2016keep,lassner2017unite,pavlakos2019expressive,ye2023decoupling}. The traditional approach estimates the model parameters by iteratively fitting the model to 2D measurements using handcrafted objectives and energy minimization techniques~\cite{bogo2016keep}. However, this optimization process contains multiple local minima, is sensitive to the choice of initialization and typically slow. To avoid those drawbacks, regression methods train a neural network to predict the human model parameters directly from images. But no existing feed-forward system achieves both accurate 3D reconstruction and image-model alignment, especially in the monocular setting. A synergy between the regression and optimization paradigms has been established~\cite{kolotouros2019learning,joo2021exemplar,kolotouros2021probabilistic}, where the regression estimate is further refined through optimization given additional observations (\eg, 2D keypoint detections). However, even in that case the optimization remains challenging, riddled with multiple local minima, while several prior terms are necessary to obtain a meaningful solution.

Diffusion models~\cite{ho2020denoising,song2020score} have recently gained a lot of attention for their ability to capture complex data distributions~\cite{dhariwal2021diffusion,rombach2022high}. These models learn the implicit prior of the underlying data distribution $\x$ by matching the gradient of the log density $\nabla_{\x} \log p(\x)$~\cite{song2020score}, also known as the score function. This learned prior can be utilized when solving inverse problems that aim to recover $\mathbf{x}$ from the observations $\mathbf{y}$ by incorporating the gradient of the log likelihood $\nabla_{\x} \log p(\y|\x)$, \textit{a.k.a} score guidance term, during sampling/denoising. The denoising process in diffusion models, characterized by its iterative nature, presents these models as a data-driven substitute for the iterative minimization employed in optimization-based techniques. Thus far, diffusion models have primarily been utilized in the generation of human motions based on text descriptions~\cite{tevet2022human,yuan2023physdiff,raab2023single}, rather than being harnessed as a tool for addressing inverse problems in 3D human recovery applications.

In this paper, we address this gap by leveraging diffusion models to solve inverse problems related to Human Mesh Recovery (HMR). We introduce Score-Guided Human Mesh Recovery (ScoreHMR), an approach designed to refine initial, per-frame 3D estimates obtained from off-the-shelf-regression networks~\cite{kanazawa2018end,kolotouros2019learning,kocabas2021pare,goel2023humans} based on additional observations. Our approach uses a diffusion model as a learned prior of a human body model (\eg, SMPL) parameters and guides its denoising process with a guidance term that aligns the human model with the available observation. The diffusion model, task-agnostic in nature, is trained on the generic task of capturing the distribution of plausible SMPL parameters conditioned on an input image. 
Given an initial regression estimate, we invert it to the corresponding latent of the diffusion model through DDIM~\cite{song2020denoising} inversion. Then we perform deterministic DDIM sampling with guidance, where this guidance term acts as the data term in a standard optimization setting, and the diffusion model serves as a learned parametric prior. The DDIM inversion -- DDIM guided sampling loop iterates until the body model aligns with the available observation. ScoreHMR can be conceptualized as a data-driven iterative fitting approach, achieving alignment with image observations through score guidance in the latent space of the diffusion model.

\begin{figure*}[t]
  \centering
  \includegraphics[width=\linewidth]{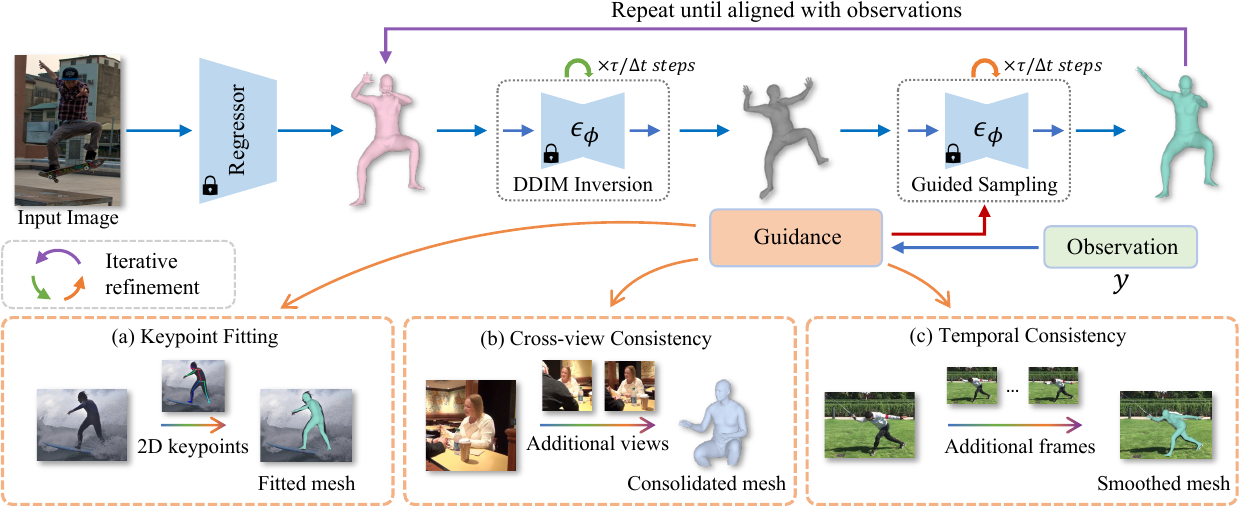}
  \caption{ 
      \textbf{Score-Guided Human Mesh Recovery and its applications.} Top row: Overview of ScoreHMR, which iteratively refines an initial regression estimate in a DDIM inversion -- DDIM guided sampling loop until the human body model aligns with the available observation. Bottom row: Applications. \textbf{(a)}: Body model fitting to 2D keypoints. \textbf{(b)}: Multi-view refinement of individual per-frame predictions with cross-view consistency guidance. \textbf{(c)}: Recovering temporally consistent and smooth 3D human motion from a video sequence given initial per-frame estimates.
  }
  \label{fig:framework}
  \vspace{-3mm}
\end{figure*}

The diffusion model can be used in many downstream applications without any need for task-specific retraining. For instance, by incorporating guidance with a keypoint reprojection term, we align the human body model with 2D keypoint detections. In scenarios with multiple uncalibrated views of a person, we employ cross-view consistency guidance to recover a 3D human mesh that maintains consistency across all viewpoints. Furthermore, in the context of inferring human motion from a video sequence, temporal consistency guidance, and optionally keypoint reprojection guidance, refines per-frame regression estimates, resulting in temporally consistent human motions. A visual summary of ScoreHMR and its applications is provided in Figure~\ref{fig:framework}.

We contribute ScoreHMR, a novel approach addressing inverse problems in 3D human recovery. We demonstrate the effectiveness of ScoreHMR with extensive experiments on the three inverse problems, refining an initial regression estimate with monocular images, multi-view images and video frames as input. Notably, our method surpasses existing optimization approaches across all datasets and evaluation settings without relying on task-specific designs or training. Beyond achieving superior results, ScoreHMR stands out as the only approach enhancing the 3D pose performance of the state-of-the-art monocular feed-forward system~\cite{goel2023humans} in the single-frame model fitting setting. We make our code and models available to support future work. We provide qualitative results on video sequences on the \href{https://statho.github.io/ScoreHMR}{project page}.

\section{Related Work}

\noindent\textbf{Regression for human mesh recovery.}
When learning to recover the 3D shape of articulated objects~\cite{yang2022banmo,stathopoulos2023learning,wu2023magicpony}, most approaches have to simultaneously learn a representation for the shape. This is not the case for the human category, since parametric models~\cite{loper2015smpl,xu2020ghum} of the human body exist, and most approaches in this paradigm learn to regress their parameters. HMR~\cite{kanazawa2018end} uses MLP layers on top of image features from a CNN to regress the SMPL model~\cite{loper2015smpl} parameters and is the canonical example in this category. Subsequent research~\cite{guler2019holopose,georgakis2020hierarchical,li2021hybrik,li2023niki,zhang2021pymaf,kocabas2021pare,li2022cliff,fang2023learning,wang2023refit} has led to many improvements in the original method. Notably, PyMAF~\cite{zhang2021pymaf} proposes a more specialized design for the CNN backbone and incorporates a mesh alignment module for SMPL parameter regression. PARE~\cite{kocabas2021pare} learns distinct features for the pose and shape parameters of SMPL and introduces a body-part-guided attention mechanism to handle occlusions. Recently, HMR 2.0~\cite{goel2023humans} proposes a fully ``transformerized" version of HMR and can effectively reconstruct unusual poses that have been difficult for previous methods. Another line of work~\cite{kolotouros2019convolutional,lin2021end,cho2022cross,lin2021mesh}, makes non-parametric predictions by directly regressing the vertices of the SMPL model. The SMPL parameters can be regressed from non-parametric predictions with an MLP without any loss in reconstruction performance~\cite{kolotouros2019convolutional}. In this work, we assume that an initial estimate in the form of SMPL parameters from a regression network is available and our goal is to improve it with our proposed approach.

\noindent\textbf{Optimization for human mesh recovery.}
Methods falling under this category~\cite{bogo2016keep,lassner2017unite,xiang2019monocular,pavlakos2019expressive,song2020human,ye2023decoupling} utilize iterative optimization to estimate the parameters of a human model~\cite{loper2015smpl,pavlakos2019expressive,xu2020ghum}. The objective is often formulated as an energy minimization problem by fitting a parametric model to the available observations, and consists of data and prior terms. The data terms measure the deviation between the estimated and detected features, while the prior terms impose constraints on the model parameters. Parametric priors are important during the optimization in order to obtain a meaningful solution, and several works have proposed a variety of them~\cite{bogo2016keep,pavlakos2019expressive,rempe2021humor,kolotouros2021probabilistic,davydov2022adversarial,tiwari2022pose}.

Nonetheless, optimization suffers from many difficulties, including sensitivity to parameter initialization, the existence of multiple local minima and the trade-off between the data and prior terms. Regression methods often serve as an initial point for an optimization-based method, which refines the estimated parameters until a convergence criterion is met~\cite{kolotouros2019learning,joo2021exemplar,kolotouros2021probabilistic}. This practice not only makes the optimization converge faster, but also typically results in a better solution since a lot of local minima are avoided. The need for multi-stage optimization procedures, as followed by early systems (\eg, SMPLify~\cite{bogo2016keep}), is also alleviated since the regressed parameters are typically close to a good solution. We evaluate our proposed approach in this setting, where our aim is to refine an initial regression estimate.  

\noindent\textbf{Solving inverse problems with diffusion models.}
Diffusion models~\cite{sohl2015deep,ho2020denoising,song2020score} are used to represent complex distributions, exhibiting remarkable success in various applications such as text-to-image generation~\cite{rombach2022high,saharia2022photorealistic}, personalization~\cite{ruiz2023dreambooth,han2023svdiff}, image editing~\cite{han2024proxedit} and video inpainting~\cite{zhang2023avid}. Their state-of-the-art performance in image generation~\cite{dhariwal2021diffusion} has led to their usage as structural priors when solving inverse problems in image processing applications, such as image inpainting~\cite{chung2022improving,chung2022diffusion,song2023pseudoinverse}, super-resolution~\cite{song2023pseudoinverse}, deblurring~\cite{chung2022diffusion} and colorization~\cite{chung2022improving} among others. Diffusion models have not been used to solve inverse problems in the context of 3D human pose and shape estimation, and our work aims to bridge this gap.
\section{Background}
\label{sec:background}

\noindent\textbf{Diffusion models.}
We first offer some background for diffusion models, namely the denoising diffusion probabilistic model (DDPM)~\cite{ho2020denoising} formulation. Let $\xzero \sim p_{data} (\x)$ denote samples from the data distribution. Diffusion models progressively perturb data to noise -- \textit{forward process} -- via Gaussian kernels for $T$ timesteps, creating latents $\{ \xt \}_{t=1}^{T}$. The noise is added with a predefined variance schedule $\{ \zeta_{t} \}_{t=1}^{T}$, such that we obtain a standard Gaussian distribution when $t=T$, \ie $\xT \sim \gauss$. Latents $\xt$ can be directly sampled from a data point $\xzero$ as $q(\xt | \xzero) = \mathcal{N} (\sqrt{\alpha_t} \xzero, (1-\alpha_{t}) \mathbf{I})$, where $\alpha_{t} := \prod_{s=1}^{t} (1 - \zeta_{s})$. A denoising model $\dm$ is trained to predict the added noise to a clean sample via minimization of the following re-weighted evidence lower bound~\cite{kingma2013auto,ho2020denoising}:
\begin{equation}
    \mathcal{L}_{simple} (\mathbf{\phi}) = \mathbb{E}_{\xzero,t,\eps} ||\dm (\xt, t) -  \eps||^{2},
\end{equation}
where $t$ is sampled uniformly from $\{1, .., T\}$, and noise $\eps$ is added to a clean sample $\xzero \sim p_{data}$ to get a noisy sample $\xt$. Once the denoising model $\dm$ is learned, we can use it to generate samples from the diffusion model by sampling $\xT \sim \gauss$ and iteratively refining it with $\dm$. The predicted noise for a latent $\xt$ at timestep $t$ (noise level) from the denoising model $\dm$ is related to the score of the model at that timestep~\cite{song2020score}:
\begin{equation}
    \label{eq:noise_score}
    \dm (\xt, t) = - \sqrt{1 - \alpha_{t}}  \nabla_{\xt} \log p (\xt).
\end{equation}

Since the sampling process -- \textit{reverse process} -- of the DDPM formulation is known to be slow~\cite{ho2020denoising,song2020denoising}, Song \etal~\cite{song2020denoising} proposed the denoising diffusion implicit model (DDIM) formulation for diffusion models, which defines the diffusion process as a non-Markovian process with the same forward marginals as DDPM. This enables faster sampling with the sampling steps given by:
\begin{equation}
    \label{eq:ddim_sampling}
    \small
    \x_{t-1} = \sqrt{\alpha_{t-1}} \xzt + \sqrt{1 - \alpha_{t-1} - \sigma_{t}^{2}}\dm (\xt, t) + \sigma_{t} \mathbf{z},
\end{equation}
where $\mathbf{z} \sim \gauss$, $\sigma_{t}$ is the variance of the noise used during sampling, and $\xzt$ denotes the predicted $\xzero$ from $\xt$ and is given by:
\begin{align}
    \label{eq:pred_x_start}
    \xzt &= \frac{1}{\sqrt{\alpha_{t}}} (\xt - \sqrt{1 - \alpha_{t}} \dm (\xt, t)),\\ \nonumber
         &\simeq \frac{1}{\sqrt{\alpha_{t}}} (\xt + (1 - \alpha_{t}) \nabla_{\xt} \log p(\xt)).
\end{align}

By setting  $\sigma_{t}$ to 0, the sampling process becomes deterministic and enables inversion of samples from $p_{data}$ to their corresponding latents~\cite{song2020denoising}. The same framework can be used for modeling conditional distributions, by incorporating the conditional information in the forward and reverse processes~\cite{dhariwal2021diffusion}. 
\section{Method}

\noindent\textbf{Body model.}
SMPL~\cite{loper2015smpl} is a parametric human body model. It consists of pose $\thb \in \mathbb{R}^{24 \times 3}$ and shape $\bb \in \mathbb{R}^{10}$ parameters, and defines a mapping $\mathcal{M}(\thb, \bb)$ from the human body parameters to a body mesh $M \in \mathbb{R}^{N \times 3}$, where $N = 6980$ is the number of mesh vertices. For a given output mesh $M$, the 3D body joints $J$ can be computed as a linear combination of the mesh vertices $J = W M$, where $W$ is a pre-trained linear regressor.

\vspace{1mm}
\noindent\textbf{Problem statement.}
Suppose we have observations $\y \in \mathbb{R}^{n}$ that relate to some unknown signal $\xzero \in \mathbb{R}^{m}$ through:
\begin{equation}
    \label{eq:inv_p}
    \y = \mathcal{A}(\xzero) + \mathbf{\eta},
\end{equation}
where $\mathcal{A}(\cdot)$ is a forward operator and $\mathbf{\eta}$ is the observation noise. Our goal is to recover $\xzero$ from $\y$, \ie solve the inverse problem. We are interested in recovering the SMPL parameters $\xzero = \{ \thb_{0}, \bb_{0} \}$ from observations $\y$ (\eg, 2D keypoint detections), from which the closed-form map to $\xzero$ is intractable. Solutions to this family of problems are given through iterative optimization by minimization:
\begin{equation}
    \argmin_{\xzero} = \mathcal{L}_{data}(\xzero) + \mathcal{L}_{prior}(\xzero),
\end{equation}
where $\mathcal{L}_{data}$ measures the deviation between the estimated and detected features and $\mathcal{L}$ consists of several prior terms necessary to obtain a plausible solution.

In our setting, we are given an input image $I$ of a person and the corresponding SMPL estimate $\x_{reg} = \{ \thb_{reg}, \bb_{reg} \}$ from regression. Our goal is to improve $\x_{reg}$ in the presence of additional observations $\mathbf{y}$. In order to achieve this we propose an approach that injects suitable information in the denoising process of a diffusion model through the log likelihood score, as described next.

\subsection{Score-Guided Human Mesh Recovery}
\label{seq:sghmr}

\phantom{11}
Our main objective is to explore how we can leverage diffusion models to solve inverse problems for human mesh recovery applications. Here, we assume that an initial estimate $\xreg$ for the SMPL parameters is acquired through any off-the-shelf regression network such as~\cite{kanazawa2018end,kolotouros2019learning,goel2023humans}, while observations $\y$ are also automatically detected. Furthermore, we assume that we have access to a trained diffusion model $\dm (\xt, t, I)$ that sufficiently captures the conditional distribution of SMPL model parameters given an input image $I$. Our goal is to improve $\xreg$ with the help of the diffusion model and detected observations $\y$.

To use the regression estimate $\xreg$ as an initial point, we invert it to the latent $\x_\tau$ at noise level $\tau$ with the deterministic DDIM inversion process:
\begin{equation}
    \label{eq:ddim_inv}
   \x_{t+1} = \sqrt{\alpha_{t+1}} \xzt + \sqrt{1-\alpha_{t+1}} \dm (\xt, t, I).
\end{equation}
Running the deterministic DDIM sampling starting from $\x_\tau$, we would get back the initial estimate $\xreg$. We found that this reconstruction error is less than $10^{-3}$ per dimension, which suggests that the DDIM inversion -- DDIM sampling loop works as intended. However, we are not interested in getting back the initial regression estimate, but we wish to improve it based on the available observation $\y$.

Ideally, we would like to use the conditional score $\nabla_{\xt} \log p (\xt | I, \y)$ during DDIM sampling instead of the score $\nabla_{\xt} \log p (\xt | I)$ of the data distribution. Using Bayes rule we can write the score $\nabla_{\xt} \log p (\xt | I, \y) = \nabla_{\xt} \log p(\xt | I) + \nabla_{\xt} \log p(\mathbf{y} | I, \xt )$, where the first term is the score of the diffusion model $\dm (\xt, t, I)$. However, the issue with this posterior sampling approach is that there does not exist an analytical formulation for the likelihood score $\nabla_{\xt} \log p(\mathbf{y} | I, \xt )$. To resolve this, a recent line of work estimates the likelihood under some mild assumptions~\cite{chung2022diffusion,song2023pseudoinverse}. Inspired by~\cite{chung2022diffusion}, by assuming that the observation noise $\mathbf{\eta}$ in \cref{eq:inv_p} is Gaussian, we get:
\begin{equation}
\label{eq:approx}
\begin{split}
    \nabla_{\xt} \log p(\y|I, \xt) & \simeq \nabla_{\xt} \log p(\y|I, \xzt) \\
                                   & = - \rho\nabla_{\xt} ||\y - \mathcal{A}(\xzt) ||_{2}^{2},
\end{split}
\end{equation} 
where $\rho$ can be viewed as a tunable step size. Approximating the likelihood score with \cref{eq:approx}, we apply guidance to the deterministic DDIM sampling process, with the sampling equations seen below:
\begin{align}
    \label{eq:ddim_g}
    \xztt &= \frac{1}{\sqrt{\alpha_{t}}} (\xt - \sqrt{1 - \alpha_{t}} \dmh (\xt, t, I)),\\
    \x_{t-1} &= \sqrt{\alpha_{t-1}} \xztt + \sqrt{1 - \alpha_{t-1}}\dmh (\xt, t, I).\nonumber
\end{align}
where $\dmh$ is the modified noise prediction after guidance:
\begin{equation}
    \label{eq:mod_noise}
    \dmh =\dm (\xt, t, I) + \rho\sqrt{1 - \alpha_{t}}\nabla_{\xt} ||\y - \mathcal{A}(\xzt) ||_{2}^{2}. 
\end{equation}

We use DDIM inversion (\cref{eq:ddim_inv}) followed by guided DDIM sampling (\cref{eq:ddim_g,eq:mod_noise}) in a loop, aligning the human body model with the detected observations. The loop stops when the relative change of the guidance loss $\mathcal{L}_{g}=||\y - \mathcal{A}(\xzt) ||_{2}^{2}$ is below a given threshold $\lambda_{thr}$. We provide a pseudo-code implementation of ScoreHMR in Appendix~\textcolor{red}{A}.

\subsection{Model Design and Training}

\phantom{11}
Without loss of generality, we choose to model only the pose SMPL parameters with our diffusion model, \ie $\xzero = \thb$, to maintain a fair comparison with optimization methods utilizing a learned pose prior (\eg, ProHMR~\cite{kolotouros2021probabilistic}). We emphasize that the shape parameters $\bb$ of SMPL can also be accommodated using the same approach and we present results from such experiments in Appendix \textcolor{red}{F}. However, we do not notice any performance improvement by including the SMPL $\bb$ in ScoreHMR. One plausible explanation is that inferring $\bb$ from a single image is relatively more straightforward compared to inferring $\thb$ for existing methods~\cite{kanazawa2018end,kolotouros2019learning,goel2023humans}.

In our setting, we are given an input image $I$ of a person, which we encode with a CNN backbone $g$ and obtain a context feature $\mathbf{c} = g (I)$. We model the distribution of plausible poses for that person conditioned on $I$ with a diffusion model $\dm(\xt, t, c=g(I))$. The backbone $g$ can be either trained end-to-end with $\dm$ or remain frozen while training the diffusion model. In the latter case, we can use the features from the backbone of a regression network~\cite{kanazawa2018end,kolotouros2019learning,kocabas2021pare,kolotouros2021probabilistic}. We did not observe any performance improvement from training $g$ end-to-end with $\dm$, and therefore, we acquire the context feature $\mathbf{c}$ from a pretrained regression network in all of our experiments.

\vspace{1mm}
\noindent\textbf{Architecture.}
We follow~\cite{zhou2019continuity} and use the 6D representation for 3D rotations, thus $\xzero$ is a 144-dimensional vector. The denoising model $\dm$ is comprised of 3 MLP blocks that are conditioned on the timestep $t$ and image features $\mathbf{c}$. The model is given a noisy sample $\xt$ for the pose parameters, the timestep $t$ and image features $\mathbf{c}$ as input. First, we use a linear layer to project $\xt$ to the features $\mathbf{h}^{(1)}$ given as input to the first MLP block. We condition the input features $\mathbf{h}^{(i)} \in \mathbb{R}^{144}$ of each MLP block on the timestep $t$, by applying scaling and shifting to get the features $\mathbf{h}_{t}^{(i)} = \mathbf{t}_{s} \mathbf{h}^{(i)} + \mathbf{t}_{b}$, where $(\mathbf{t}_{s}, \mathbf{t}_{b}) \in \mathbb{R}^{2 \times 144} = MLP(\psi (t) )$ is the output of a MLP with a sinusoidal encoding function $\psi$. Then, we condition each MLP block on the image features by concatenating $\mathbf{h}_{t}^{(i)}$ and $\mathbf{c}$. Additional details are provided in Appendix~\textcolor{red}{B}.

\vspace{1mm}
\noindent\textbf{Training.}
Let us assume that we have a collection of images paired with SMPL pose annotations. Then, we could train the diffusion model with its standard training loss:
\begin{equation}
    \mathcal{L}_{DM} (\mathbf{\phi}) = \mathbb{E}_{(I, \x_{0}),t,\eps} ||\dm (\xt, t, I) -  \eps||^{2}.
\end{equation}
Unfortunately, such paired annotations are not generally available, so we use pseudo ground-truth SMPL pose annotations from various datasets (see~\cref{seq:setup}).

\subsection{Applications of ScoreHMR}

\phantom{11}
In this part we show how we can use our approach for solving HMR-related inverse problems. We highlight that for all these applications we use the same trained diffusion model with no per-task training.

\vspace{1mm}
\noindent\textbf{Body model fitting.}
In this setting the detected image observations are 2D keypoints detections $\y_{kp}$ and their confidences $\y_{conf}$. Optimization approaches fit the SMPL body model to the 2D keypoints by minimizing $\lambda_{J} E_{J} + \lambda_{prior} E_{prior}$, where $E_{J}$ penalizes the deviations between the projected model joints and the detected joints and $E_{prior}$ include prior energy terms for the pose and shape parameters of SMPL.

Typically the predicted weak-perspective camera from a regression network is converted to a perspective camera $\pi = (R, \gamma)$ based on the bounding box of a person and is also included as a variable to be optimized. The camera $\pi$ has fixed focal length and intrinsics $K$. Since the parameters $\thb$ already include a global orientation, $R \in \mathbb{R}^{3 \times 3}$ is assumed to be identity and only the camera translation $\mathbf{\gamma} \in \mathbb{R}^{3}$ is optimized along with the human body model parameters.

In this setting, the forward operator that relates the body model parameters with the detected joints is $\Pi_{K}(W \mathcal{M} (\xzero, \bb) + \mathbf{\gamma})$, where $\Pi_{K}$ is the projection matrix with camera intrinsics $K$ and $W$ is a matrix that regresses the 3D model joints from the mesh vertices of the model. This means that the guidance loss in \cref{eq:mod_noise} becomes:
\begin{equation}
    \label{eq:kp_repr}
    \mathcal{L}_{repr} = \y_{conf} || \Pi_{K}(W \mathcal{M} (\xzt, \bb) + \mathbf{\gamma}) - \y_{kp}||_{2}^{2}.
\end{equation}
The camera translation $\mathbf{\gamma}$ is also optimized with $\mathcal{L}_{repr}$ as in standard optimization procedures. 

\vspace{1mm}
\noindent\textbf{Multi-view refinement.}
In this setting we have a set $\{ I^{(n)} \}_{n=1}^{N}$ of uncalibrated views of the same person, and their monocular regression estimate that we want to improve based on information from the other views. For each frame, we decompose the pose parameters $\xzero^{(n)}$ to global orientation $\x_{0,gl}^{(n)}$ and body pose parameters $\x_{0,b}^{(n)}$. We can consolidate all single-frame predictions to improve $\x_{0,b}^{(n)}$ with a cross-view consistency guidance loss:
\begin{equation}
    \label{eq:multi_view}
    \mathcal{L}_{MV} = \sum_{n=1}^{N}|| \hat{\x}_{0,b}^{(n)}(\x_{t}^{(n)}) - \Bar{\x}_{0,b} ||_{2}^{2},
\end{equation}
where $\Bar{\x}_{0,b} = \frac{1}{N} \sum_{n}^{N} \x_{0,b}^{(n)}(\x_{t}^{(n)})$ and its minimization is equivalent to minimizing the squared distance between all pairs of body poses.

\vspace{1mm}
\noindent\textbf{Human motion refinement.}
Although our model has been trained in the monocular setting, we can use the learned conditional distribution to obtain temporally consistent and smooth predictions in a video sequence $V = \{ I^{(n)} \}_{n=1}^{N}$. In this setting, the forward operator is the identity function and the observations are the pose predictions of the previous frame in the sequence. We can enforce temporal consistency with the following guidance loss:
\begin{equation}
    \label{eq:smoothness}
    \mathcal{L}_{temp} = \sum_{n=2}^{N} || \hat{\x}_{0}^{(n)}(\xt) - \hat{\x}_{0}^{(n-1)}(\xt)||_{2}^{2}.
\end{equation}

Guidance with the previous loss can be considered as a learnable smoothing operation that makes sure that the smoothed parameters remain consistent with the image evidence under the image-conditional distribution captured by the diffusion model. We can optionally use additional guidance with the keypoint reprojection loss in \cref{eq:kp_repr} when 2D keypoint detections are available.
\section{Experiments}

\vspace{1mm}
\noindent\textbf{Training.}
\label{seq:setup}
We use the typical datasets for training, \ie, Human3.6M~\cite{ionescu2013human3}, MPI-INF-3DHP~\cite{mpi-inf}, COCO~\cite{coco} and MPII~\cite{mpii}. The quality of the pseudo ground-truth pose annotations plays an important role for training the diffusion model. We compare two models trained with pseudo ground-truth from SPIN~\cite{kolotouros2019learning} and EFT~\cite{joo2021exemplar} respectively. To showcase that ScoreHMR can work with image features from various HMR models, we also train two different versions of $\dm$ with image features from ProHMR~\cite{kolotouros2021probabilistic} and PARE~\cite{kocabas2021pare} respectively. When training with PARE features, we only use its pose features. Implementation details and hyper-parameters are provided in Appendix~\textcolor{red}{B}.

\vspace{1mm}
\noindent\textbf{Evaluation datasets.}
For the body model fitting to 2D keypoints and human motion refinement settings, we conduct evaluation on the test set of 3DPW~\cite{von2018recovering} and on the split of EMDB~\cite{kaufmann2023emdb} that contains the most challenging sequences (\ie, EMDB 1). For the multi-view refinement experiment, we report results on Human3.6M~\cite{ionescu2013human3} and Mannequin Challenge~\cite{li2019learning}. For Mannequin Challenge we use the annotations produced by Leroy \etal~\cite{leroy2020smply} and employ the entire dataset for evaluation.

\vspace{1mm}
\noindent\textbf{Evaluation setup.}
In order to demonstrate the efficacy of our approach in refining the regression estimates from various networks and accuracy levels, we use the predictions from the less accurate ProHMR's regression network~\cite{kolotouros2021probabilistic} and the highly accurate HMR~2.0~\cite{goel2023humans} as our starting points. For experiments with HMR~2.0, we use the HMR~2.0b model, which trains longer and on more data than HMR~2.0a, and can reconstruct humans in challenging and unusual poses.

\begin{table}
	\centering
    \small
	\hspace{-3mm}
	\begin{tabular}{ @{}lcccc@{} }
		\toprule
		& Features & Fits & 3DPW (14) & EMDB 1 (24)\\
        \midrule
        ProHMR~\cite{kolotouros2021probabilistic} &-&-&59.8&86.1\\
        \phantom{1} + ScoreHMR &ProHMR&SPIN&55.7&77.8\\
        \phantom{1} + ScoreHMR &ProHMR&EFT&55.5&77.4\\
        \phantom{1} + ScoreHMR &PARE&SPIN&55.6&77.4\\
        \phantom{1} + ScoreHMR &PARE&EFT&54.7&77.1\\
        \midrule
        HMR 2.0~\cite{goel2023humans} &-&-&54.3&78.7\\
        \phantom{1} + ScoreHMR &ProHMR&SPIN&52.4&76.5\\
        \phantom{1} + ScoreHMR &ProHMR&EFT&51.3&\textbf{76.4}\\
        \phantom{1} + ScoreHMR &PARE&SPIN&52.4&76.6\\
        \phantom{1} + ScoreHMR &PARE&EFT&\textbf{51.1}&76.6\\
		\bottomrule
	\end{tabular}
	\vspace{-2mm}
	\caption{\textbf{Ablation study.} ScoreHMR is initialized by the corresponding regression results. All numbers are PA-MPJPE in mm. Parenthesis denotes the number of body joints used to compute PA-MPJPE.}
	\label{tab:ablation}
\end{table}

\begin{table}
	\centering
        \small
	\hspace{-3mm}
	\begin{tabular}{ @{}lcc@{} }
		\toprule
		& 3DPW (14) & EMDB 1 (24)\\
		\midrule
	    LGD~\cite{song2020human} &55.9&81.1\\	
        LFMM~\cite{choutas2022learning} &52.2&-\\
        \midrule
        ProHMR~\cite{kolotouros2021probabilistic}
            &59.8&86.1\\
        \phantom{1} + SMPLify~\cite{bogo2016keep}
            &60.9&84.6\\
        \phantom{1} + fitting~\cite{kolotouros2021probabilistic}
            &55.1&79.8\\
        \phantom{1} + ScoreHMR-a
            &55.7&77.8\\
        \phantom{1} + ScoreHMR-b
            &54.7&77.1\\
        \midrule
        HMR 2.0~\cite{goel2023humans}
            &54.3&78.7\\
        \phantom{1} + SMPLify~\cite{bogo2016keep}
            &60.1&83.5\\
        \phantom{1} + fitting~\cite{kolotouros2021probabilistic}
            &55.1&80.1\\
        \phantom{1} + ScoreHMR-a
            &52.4&\textbf{76.5}\\
        \phantom{1} + ScoreHMR-b
            &\textbf{51.1}&76.6\\
		\bottomrule
	\end{tabular}
	\vspace{-2mm}
	\caption{\textbf{Evaluation of different model fitting methods.} The fitting algorithms are initialized by the corresponding regression results, except LGD~\cite{song2020human} and LFMM~\cite{choutas2022learning}. All numbers are PA-MPJPE in mm. Parenthesis denotes the number of body joints used to compute PA-MPJPE.}
	\vspace{-4mm}
	\label{tab:modelfitting}
\end{table}

\subsection{Quantitative Evaluation}

\subsubsection{Body model fitting}
We evaluate the accuracy of methods that fit the SMPL body model to 2D keypoint detections. The keypoints are detected with OpenPose~\cite{cao1812openpose}.

\vspace{1mm}
\noindent\textbf{Ablation study.}
First, we provide an ablation study of the core components of ScoreHMR. We benchmark ScoreHMR with diffusion models trained with frozen image features from ProHMR~\cite{kolotouros2021probabilistic} and PARE~\cite{kocabas2021pare}, and pseudo ground-truth pose annotations from SPIN~\cite{kolotouros2019learning} and EFT~\cite{joo2021exemplar}. We report results of iterative refinement with ScoreHMR using the keypoint reprojection loss $\mathcal{L}_{repr}$ in \cref{eq:kp_repr}. Following the typical protocols of prior work~\cite{kolotouros2019learning,kolotouros2021probabilistic} we use the PA-MPJPE metric for evaluation and present results in Table~\ref{tab:ablation}. From Table~\ref{tab:ablation} we observe that running ScoreHMR on top of regression reduces the 3D pose errors in all cases. We also observe that iterative refinement with ScoreHMR is robust to the choice of image features and pseudo ground-truth. The diffusion model, trained with PARE image features and fits from EFT, attains the highest performance. We use ScoreHMR with our worst (ProHMR features \& SPIN fits) and best (PARE features \& EFT fits) models for evaluation in the rest of the paper, denoting them as ScoreHMR-a and ScoreHMR-b respectively.

\noindent\textbf{Comparison with optimization methods.}
Next, we compare with model fitting baselines that are trained to optimize starting from the canonical pose and shape (\ie, LGD~\cite{song2020human}, LFMM~\cite{choutas2022learning}) as well as with baselines that can use the parameters from a regression network as a starting point (\ie, SMPLify~\cite{bogo2016keep}, ProHMR-fitting~\cite{kolotouros2021probabilistic}). We benchmark SMPLify (single-stage implementation from~\cite{kolotouros2019learning}) and ProHMR-fitting starting from the predictions of the ProHMR's regression network~\cite{kolotouros2021probabilistic} and those of HMR~2.0~\cite{goel2023humans}. Results are reported in Table~\ref{tab:modelfitting}. Performing SMPLify on top of regression increases the 3D pose errors, while ProHMR-fitting fails to improve the performance of HMR~2.0. Iterative refinement with ScoreHMR reduces the 3D pose errors in all cases, and ScoreHMR-b outperforms all baselines.

\begin{table}
	\centering
	\footnotesize
	\hspace{-3mm}
    \resizebox{\columnwidth}{!}{
    	\begin{tabular}{ @{}lcccc@{} }
    		\toprule
    		& \multicolumn{2}{c}{H36M (14)}  & \multicolumn{2}{c}{Mannequin (17)}\\
            \cmidrule(lr){2-3}\cmidrule(lr){4-5}
    		& MPJPE $\downarrow$ & PA-MPJPE $\downarrow$ & MPJPE $\downarrow$ & PA-MPJPE $\downarrow$\\
    		\midrule
            ProHMR~\cite{kolotouros2021probabilistic}
                &65.1&43.7&165.3&86.8\\
            \phantom{1} + fitting~\cite{kolotouros2021probabilistic}
                &59.6&34.5&162.6&80.2\\
            \phantom{1} + ScoreHMR-a
                &55.8&34.1&162.0&81.1\\
            \phantom{1} + ScoreHMR-b
                &51.9&34.2&157.7&80.2\\
            \midrule
            HMR 2.0~\cite{goel2023humans}
                &52.8&35.6&156.0&90.1\\
            \phantom{1} + fitting~\cite{kolotouros2021probabilistic}
                &52.6&32.9&155.5&79.4\\
            \phantom{1} + ScoreHMR-a
                &47.9&\textbf{28.4}&151.0&79.3\\
            \phantom{1} + ScoreHMR-b
                &\textbf{44.7}&29.0&\textbf{148.3}&\textbf{79.1}\\
            \bottomrule
    	\end{tabular}
    }
	\vspace{-2mm}
	\caption{\textbf{Evaluation of multi-view refinement.} We compare our proposed approach with the single-view 3D reconstruction and an optimization-based method~\cite{kolotouros2021probabilistic}. Parenthesis denotes the number of body joints used to compute MPJPE and PA-MPJPE.}
	\vspace{-4mm}
	\label{tab:multiview}
\end{table}

\subsubsection{Multi-view refinement}
Next, we evaluate the capability of ScoreHMR at refining the per-view regression estimates when several uncalibrated views of the same person are available. For this task, we use guidance with the cross-view consistency loss $\mathcal{L}_{MV}$ in \cref{eq:multi_view}. We test our approach on the Human3.6M~\cite{ionescu2013human3} and the Mannequin Challenge~\cite{li2019learning} (some YouTube videos were missing) datasets, reporting MPJPE and PA-MPJPE following~\cite{kolotouros2021probabilistic}. We compare with the individual per-view regression predictions as well as with an optimization-based method~\cite{kolotouros2021probabilistic}. Results are shown in Table~\ref{tab:multiview}.  Results from Table~\ref{tab:multiview} show that both ScoreHMR and ProHMR-fitting improve the per-frame predictions, but our approach consistently leads to lower MPJPE errors. This happens because refining the body poses at a given noise level also influences the global orientation in the next noise level of the diffusion model, as the model captures the joint distribution of SMPL poses $\thb$. This is not possible with ProHMR-fitting, since only the body poses are updated during the optimization process. Notably, the runtime of ScoreHMR is remarkably swift, requiring only 1.5 minutes for the entire Mannequin Challenge dataset, which contains 20K frames.

\subsubsection{Human motion refinement}
In this part, we evaluate ScoreHMR at refining the single-frame regression estimates in a video sequence with 2D keypoint detections. In this setting, we use guidance with $\mathcal{L}_{repr}$ and $\mathcal{L}_{temp}$ terms. Following prior work~\cite{kanazawa2019learning} we also report the acceleration error ($mm/s^{2}$), computed as the difference in acceleration between the ground-truth and predicted 3D joints.  We use all SMPL body joints for computing this error in EMDB 1, in contrast to the evaluation in~\cite{kaufmann2023emdb} that uses specific joints for some temporal metrics (\eg Jitter).

\begin{table}
	\centering
	\footnotesize
	\hspace{-3mm}
    \resizebox{\columnwidth}{!}{
        \begin{tabular}{ @{}lcccc@{} }
            \toprule
            & \multicolumn{2}{c}{3DPW (14)} & \multicolumn{2}{c}{EMDB 1 (24)}\\
            \cmidrule(lr){2-3}\cmidrule(lr){4-5}
            & PA-MPJPE $\downarrow$ & Acc Err $\downarrow$ & PA-MPJPE $\downarrow$ & Acc Err$\downarrow$\\
    	    \midrule
            Vibe~\cite{kocabas2020vibe}
                &56.7&31.5&85.7&43.8\\
            Vibe-opt~\cite{kocabas2020vibe}
                &63.9&42.1&83.6&41.4\\
            \midrule
            ProHMR~\cite{kolotouros2021probabilistic}
                &59.8&25.0&86.1&37.7\\
            \phantom{1} + fitting~\cite{kolotouros2021probabilistic}
                &54.5&14.0&77.9&18.4\\
            \phantom{1} + ScoreHMR-a
                &54.9&11.4&76.5&12.8\\
            \phantom{1} + ScoreHMR-b
                &53.9&11.2&75.7&12.1\\
            \midrule
            HMR 2.0~\cite{goel2023humans}
                &54.3&17.3&78.7&23.7\\
            \phantom{1} + fitting~\cite{kolotouros2021probabilistic}
                &53.8&14.1&76.2&20.0\\
            \phantom{1} + ScoreHMR-a
                &51.7&\textbf{10.7}&\textbf{75.1}&11.9\\
            \phantom{1} + ScoreHMR-b
                &\textbf{50.5}&11.1&75.3&\textbf{11.9}\\
            \bottomrule
    	\end{tabular}
    }
	\vspace{-2mm}
	\caption{\textbf{Evaluation of human motion refinement.} We compare different model fitting algorithms and our proposed approach in a temporal setting. Parenthesis denotes the number of body joints used to compute PA-MPJPE and Acc Err.}
	\label{tab:temp_fitting}
\end{table}

\begin{figure}[t]
    \centering
    \includegraphics[width=\columnwidth]{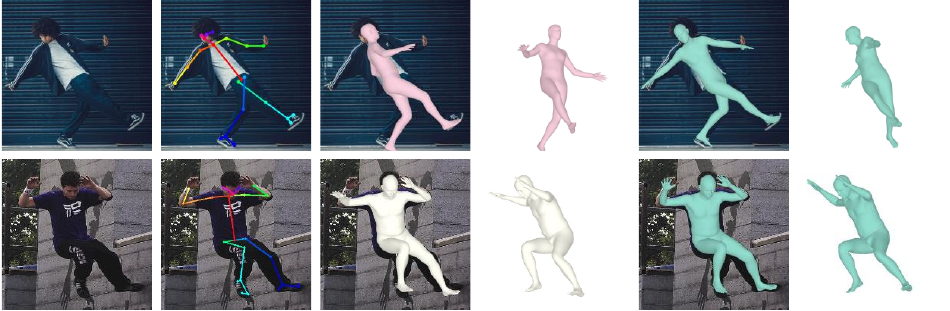}
    \caption{
        \textbf{Qualitative evaluation of ScoreHMR} Pink: Regression with ProHMR~\cite{kolotouros2021probabilistic}. White: Regression with HMR 2.0~\cite{goel2023humans}. Green: Regression + ScoreHMR (ours).
    }
	\vspace{-3mm}
  \label{fig:fit}
\end{figure}

\begin{figure*}[t]
  \centering
  \includegraphics[width=\textwidth]{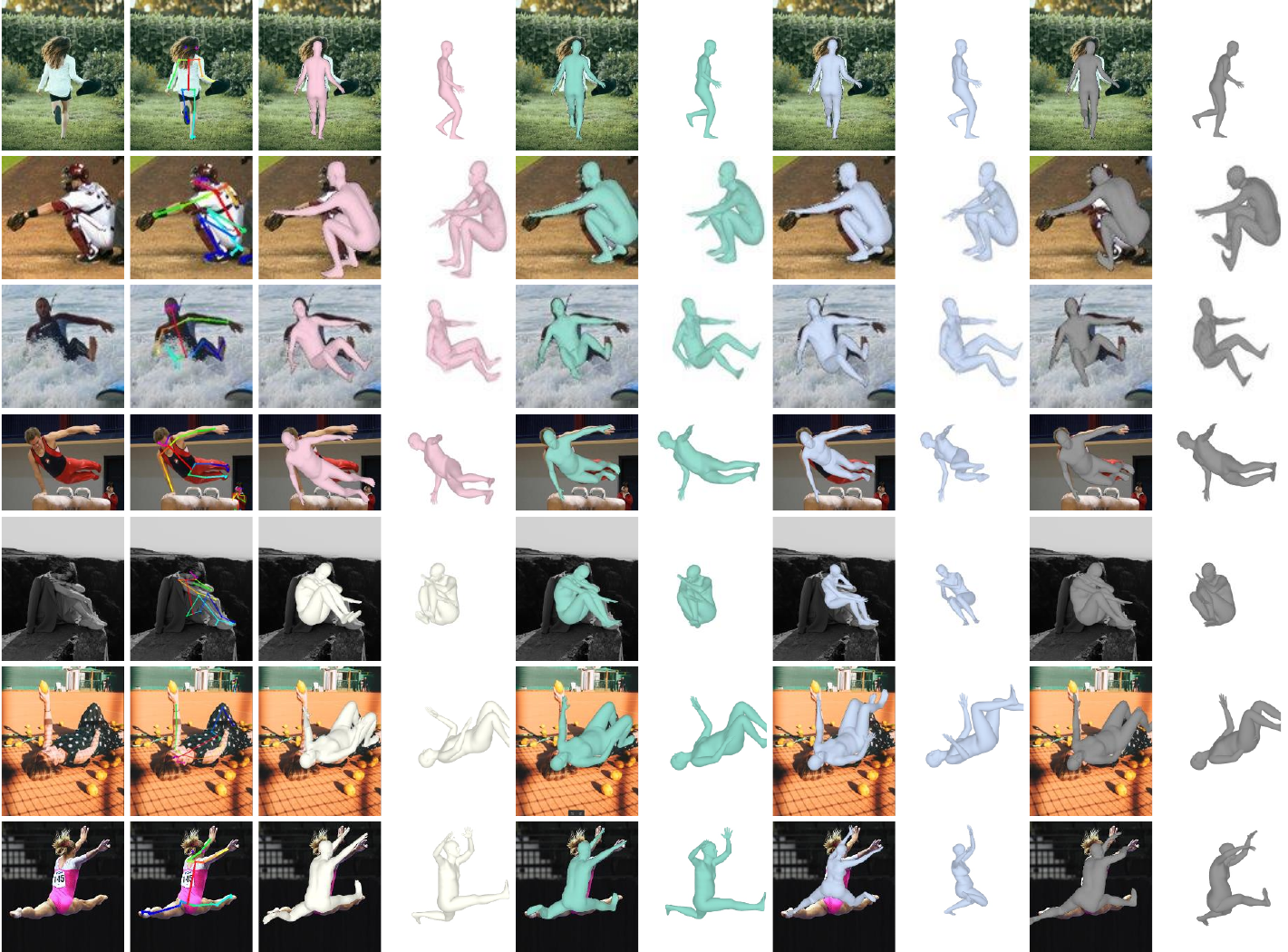}
  \caption{
      \textbf{Body model fitting results.} Pink: Regression (ProHMR~\cite{kolotouros2021probabilistic}). White: Regression (HMR 2.0~\cite{goel2023humans}). Green: Regression + ScoreHMR (ours). Blue: Regression + ProHMR-fitting~\cite{kolotouros2021probabilistic}. Grey: Regression + SMPLify~\cite{bogo2016keep}.
  }
  \label{fig:fit_comp}
\end{figure*}

We compare our approach with the temporal mesh optimization baselines (VIBE-opt~\cite{kocabas2020vibe}, ProHMR-fitting~\cite{kolotouros2021probabilistic}). VIBE-opt is initialized by the temporal mesh regression result of VIBE~\cite{kocabas2020vibe}. We run ProHMR-fitting~\cite{kolotouros2021probabilistic} with the default hyperparameters adding a smoothness regularization term. We report results in Table~\ref{tab:temp_fitting}. Our approach consistently outperforms all baselines across all datasets and metrics. Notably, ScoreHMR significantly enhances temporal consistency compared to prior works, resulting in a relative improvement of $21.3\%$ (3DPW) and $40.5\%$ (EMDB~1) in acceleration error compared to ProHMR-fitting, when both methods start from the monocular regression estimate of HMR~2.0. Finally, ScoreHMR exhibits exceptional runtime efficiency requiring only 14 minutes for the entire 3DPW test set, which contains 35K frames.

\subsection{Qualitative Results}
\phantom{11}
We show qualitative results in body model fitting on top of ProHMR and HMR~2.0 regression in Figure~\ref{fig:fit}. ScoreHMR effectively aligns the body model with the detected keypoints even when the initial regression estimate is inaccurate (\eg, example of first row). Our reconstructions are valid when seen from a novel view. In addition, we compare our approach with SMPLify and ProHMR-fitting in Figure~\ref{fig:fit_comp}. Our approach achieves more faithful reconstructions than the baselines. This is more evident in challenging poses (\eg, example of last row). SMPLify encounters challenges with inaccurate keypoint detections (\eg, example of second row). ProHMR-fitting faces difficulties when there is ambiguity in the image evidence (\eg, occlusion in the example of third row). A potential cause for this issue may be the mode supervision used during ProHMR training, which leads to capturing a less diverse pose distribution as shown in~\cite{chen2023mhentropy}. We include more qualitative examples in Appendix \textcolor{red}{G}. We also encourage viewing video results on the \href{https://statho.github.io/ScoreHMR}{project page}.
\section{Conclusion}
\phantom{11}
We present ScoreHMR, an approach for solving inverse problems for 3D human pose and shape reconstruction. ScoreHMR mirrors model fitting approaches, but alignment with the image observation is achieved through score guidance in the latent space of a diffusion model. We demonstrate the effectiveness of our method with empirical results in several benchmarks and evaluation settings. ScoreHMR achieves strong performance in challenging datasets and outperforms optimization-based methods. Our work highlights the promising potential of score-guided diffusion processes as a better alternative to conventional optimization-based approaches in addressing 3D human recovery inverse problems.

{
    \small
    \bibliographystyle{ieeenat_fullname}
    \bibliography{main}
}

\begin{algorithm*}[t]
	\renewcommand{\algorithmicrequire}{\textbf{Input:}}
	\renewcommand{\algorithmicensure}{\textbf{Output:}}
	\caption{Score-Guided Human Mesh Recovery (ScoreHMR)}
     \label{algo:sghmr}
	\begin{algorithmic}[1]
		\REQUIRE Given observation $\y$, denoising model $\dm$, image features $\bc_{I}$, estimate $\x_{reg}$ from a regression network, gradient step size $\rho$, noise level $\tau$, DDIM step size $\Delta t$, threshold $\lambda_{thres}$, number of iterations for the outer refinement loop $S_{max}$.
        \FOR{$s=1$ to $S_{max}$}
            \IF{$s=1$}
                \STATE $\x_{init} \gets \xreg$ \comm{ First iteration starts with estimate from regression}
            \ELSE
                \STATE $\x_{init} \gets \xzero$ \comm{ Iteration starts with $\xzero$ from previous iteration}
            \ENDIF
            \STATE $\x_{\tau}=\text{DDIMInvert}(\x_{init}, \bc_{I})$ \comm{ Run DDIM inversion until noise level $\tau$}
		      \FOR{$t=\tau$ to $\Delta t$ with step size $\Delta t$}
                \STATE $\tilde\epsilon \gets \dm(\xt, t, \bc_{I})$ \comm{ Predict noise}
                \STATE Initialize computational graph for $\xt$
                \STATE $\hat{\x}_{0} \gets \frac{1}{\sqrt{\alpha_{t}}} (\xt - \sqrt{1 - \alpha_{t}}) \tilde\epsilon$ \comm{ Predict one-step denoised result}
                \STATE $\mathcal{L}_{g} \gets ||\y - \mathcal{A}(\hat{\x}_{0})||^{2}$ \comm{ Compute guidance loss}
                \IF {$\mathcal{L}_{g} < \lambda_{thres}$} 
                    \STATE \textbf{return} $\hat{\x}_{0}$ \comm{ Early stopping: return $\xzero$ if the loss is below a threshold}
                \ENDIF
                \STATE $\tilde\epsilon^{'} \gets \tilde\epsilon + \rho\sqrt{1 - \alpha_{t}}\nabla_{\xt}\mathcal{L}_{g}$ \comm{ Compute modified noise after score-guidance}
                \STATE $\hat{\x}_{0}^{'} \gets \frac{1}{\sqrt{\alpha_{t}}} (\xt - \sqrt{1 - \alpha_{t}}) \tilde\epsilon^{'}$ \comm{ Predict one-step denoised result with modified noise}
                \STATE $\x_{t-\Delta t} \gets {\sqrt{\alpha_{t-\Delta t}}}\hat{\x}_{0}^{'} + \sqrt{1 - \alpha_{t-\Delta t}} \tilde\epsilon^{'}$ \comm{ DDIM sampling step}
            \ENDFOR
  		\ENDFOR
		\STATE \textbf{return} $\hat{\x}_{0}^{'}$
	\end{algorithmic}  
\end{algorithm*}

\renewcommand{\thesection}{\Alph{section}}
\setcounter{section}{0}

\section{ScoreHMR Pseudo-code}
\label{sec:algo}
\phantom{11}
We illustrate a pseudo-code implementation of ScoreHMR in \cref{algo:sghmr}.

\section{Implementation Details}
\label{sec:implementation}
\phantom{11}
In this section, we provide details on the architecture and the training process of the denoising model $\dm$, and on the hyper-parameters used for guidance. Our code and pre-trained model weights are also released at \href{https://github.com/statho/ScoreHMR}{https://github.com/statho/ScoreHMR}.

\begin{figure*}[t]
  \centering
  \includegraphics[width=0.95\textwidth]{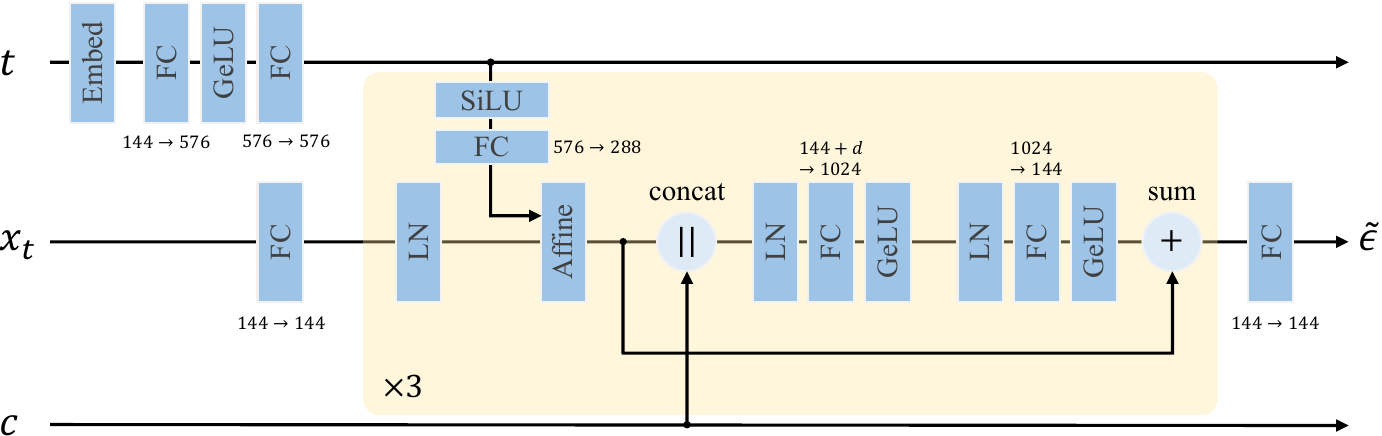}
  \caption{ \textbf{Diffusion model architecture.} Implementation of $\dm(\xt, t, \bc = g(I))$. $LN$ denotes Layer Normalization~\cite{ba2016layer}, $\parallel$ denotes concatenation, and $d$ denotes the dimension of the image features $\bc$. Rotations are parameterized with 6D representations, thus $\xzero, \xt, \tilde{\mathbf{\epsilon}}$ are 144-D vectors. 
  }
  \label{fig:arch}
\end{figure*}

\vspace{1mm}
\noindent\textbf{Denoising model architecture.}
The architecture of the denoising model $\dm$ is depicted in Figure~\ref{fig:arch}. For each trainable layer we include the number of input and output features as $d_{in} \rightarrow d_{out}$. We use image features $\bc$ from frozen HMR regression networks as discussed in the main paper. ProHMR~\cite{kolotouros2021probabilistic} uses the standard ResNet-50~\cite{he2016deep} backbone, and we use the features after the global average pooling layer, \ie the dimension of $\bc$ is 2048. PARE~\cite{kocabas2021pare} learns disentangled features for the pose and shape SMPL parameters. We only used the pose features of PARE, so $\bc$ is a $3072$-dimensional vector. 

\vspace{1mm}
\noindent\textbf{Training details.}
The total number of timesteps in the diffusion model is set to $T=1,000$ following prior work~\cite{ho2020denoising,dhariwal2021diffusion}. We use cosine variance schedule~\cite{nichol2021improved}. We train with a batch size of 128, learning rate $10^{-4}$ and Adam optimizer~\cite{kingma2014adam} for 1M iterations. We maintain an exponential moving average (EMA) copy of our model with rate of 0.995. Our implementation is in PyTorch~\cite{paszke2019pytorch}. Training takes only 6 hours on a single NVIDIA A100 GPU.

\vspace{1mm}
\noindent\textbf{Guidance details.}
The gradient step size in \cref{eq:approx} is set to $\rho_{repr} = 0.003$, $\rho_{MV} = 0.005$ and $\rho_{temp} = 30$ for $\mathcal{L}_{repr}$, $\mathcal{L}_{MV}$ and $\mathcal{L}_{temp}$ respectively. The outer refinement loop is set to $S_{max}=10$. The threshold for the early stopping criterion is set to $\lambda_{thr}=10^{-5}$. The timestep (noise level) where there refinement process starts is set to $\tau = 50$ and the DDIM step size is set to $\Delta t=2$. For multi-view refinement experiments we set $\tau = 100$ and $\Delta t=10$.

\section{Ablations}
This section provides an ablation study of the two core components of score guidance. The ablation study is performed on the 3DPW test set in the model fitting setting, starting from the regression estimate of HMR 2.0b with 54.3 PA-MPJPE. The default setting is marked with \gray{gray}. All other components are set to their default values during each component's individual ablation.

\vspace{1mm}
\noindent\textbf{Noise level $\tau$.} The Table below shows the PA-MPJPE error varying $\tau$. ScoreHMR works better for small noise levels~$t$. The one-step denoised result $\xzt$ used to compute the guidance loss (\cref{eq:mod_noise}) is more accurate for small values of $t\in [0, \tau]$.

\vspace{-5mm}
\noindent
\begin{table}[htbp]
    \centering
    \resizebox{0.82\columnwidth}{!}{
    \begin{tabular}{c|cccc}
        $\tau$ & \gray{50}&100&200&300\\ 
        \hline
        \scriptsize{HMR 2.0b + ScoreHMR} &51.1&52.3&54.3&54.5\\
    \end{tabular}
    }
    \vspace{-2mm}
\end{table}

\noindent\textbf{DDIM step size $\Delta t$.} The Table below shows the PA-MPJPE error varying the DDIM step size $\Delta t$. Even though larger DDIM step sizes result in lower PA-MPJPE in 3DPW, we find that ScoreHMR with a small step size is more robust and performs better qualitatively especially for challenging and unusual poses. A similar observation is made in~\cite{goel2023humans}, where HMR 2.0b has a higher PA-MPJPE error than HMR 2.0a, but performs better in practice.
\begin{table}[htbp]
    \centering
\resizebox{\columnwidth}{!}{
\begin{tabular}{c|cccccc}
    $\Delta t$ & \gray{2}&4&6&8&10&12\\ 
    \hline
    \scriptsize{HMR 2.0b + ScoreHMR} &51.1&49.6&48.8&48.4&48.2&48.4\\
\end{tabular}
}
\end{table}

\section{Datasets}
\phantom{11}
In this part we offer some information on the datasets used for training and evaluation. The datasets used for training are Human3.6M~\cite{ionescu2013human3}, MPI-INF-3DHP~\cite{mpi-inf}, COCO~\cite{coco} and MPII~\cite{mpii}. The datasets used for evaluation are 3DPW~\cite{von2018recovering}, EMDB~\cite{kaufmann2023emdb}, Human3.6M~\cite{ionescu2013human3} and Mannequin Challenge~\cite{li2019learning}.

\vspace{1mm}
\noindent\textbf{Human3.6M.}
It contains data for 3D human pose captured in a studio environment. Following standard practices we use subjects S1, S5, S6, S7 and S8 for training, while we use subjects S9 and S11 for evaluation in the multi-view refinement setting.

\vspace{1mm}
\noindent\textbf{MPI-INF-3DHP.}
It contains data for 3D human pose captured mainly in indoor studio environments with a marker-less setup. We use the predefined train split for training.

\vspace{1mm}
\noindent\textbf{COCO.}
It contains images in-the-wild annotated with 2D keypoints. We use this dataset only during training.

\vspace{1mm}
\noindent\textbf{MPII.}
It contains images annotated with 2D keypoints. We use this dataset only during training.

\vspace{1mm}
\noindent\textbf{3DPW.}
It is a dataset captured in indoor and outdoor locations and contains SMPL pose and shape ground-truth. We follow standard practices in the literature and only use the predefined test split for evaluation.

\vspace{1mm}
\noindent\textbf{EMDB.}
It is a new dataset captured in indoor and outdoor locations and contains SMPL pose and shape ground-truth. It includes a split (\ie, EMDB 1) with the most challenging outdoor sequences. We use EMDB 1 for evaluation.

\vspace{1mm}
\noindent\textbf{Mannequin Challenge.}
It contains videos of people staying frozen in various poses. We use the SMPL annotations from~\cite{leroy2020smply} for evaluation in this dataset.

\section{Evaluation Metrics}
\phantom{11}
Depending on the setting, we evaluate using the MPJPE, PA-MPJPE and Acc Err metrics following standard practices in the literature. The Mean Per Joint Position Error (MPJPE) computes the Euclidean error between the predicted and ground-truth 3D joints, after aligning them at the pelvis. The PA-MPJPE compute the same error after aligning the predicting the ground-truth 3D joints with Procrustes alignment. Both metrics are used for per-frame 3D human pose evaluation. The acceleration error (Acc Err) is a temporal metric that measures the average difference between ground truth 3D acceleration and predicted 3D acceleration of joints in $mm/s^{2}$.

\begin{figure*}[t]
  \centering
    \includegraphics[width=\linewidth]{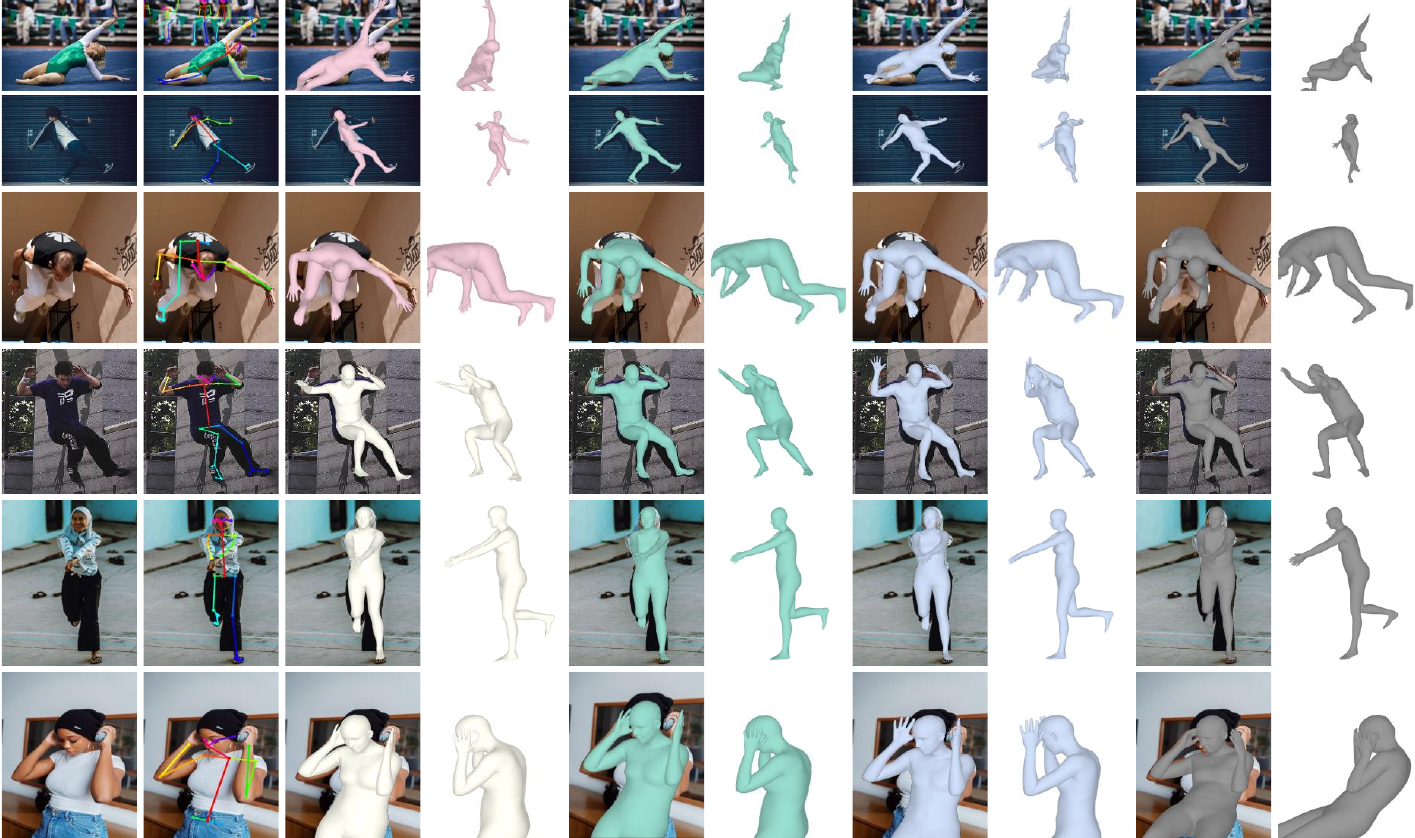}
  \caption{ \textbf{Model fitting results.} We compare our approach (green) with ProHMR-fitting (blue) and SMPLify (grey). All model fitting algorithms are initialized with regression from ProHMR (pink) or HMR 2.0b (white).
  }
  \label{fig:supp_fit}
\end{figure*}

\section{Additional Quantitative Evaluation}

\noindent\textbf{Diffusion Model for SMPL $\beta$.}
As mentioned in the main paper, we can also include the SMPL shape parameters $\bb$ in ScoreHMR. In Table~\ref{tab:dm_betas} we present results in single-frame model fitting to 2D keypoint detections, comparing ScoreHMR with and without the inclusion of SMPL $\bb$. We observe that modeling and optimizing $\bb$ with our proposed approach works well, but does not bring any significant performance improvement compared to modeling only the SMPL pose parameters $\thb$.

\begin{table}[t]
    \centering
    \resizebox{0.85\columnwidth}{!}{
        \begin{tabular} {@{}lcc@{} }
            \toprule
            & 3DPW (14) & EMDB 1 (24)\\
            \midrule
            HMR 2.0b~\cite{goel2023humans} & 54.3 & 78.7\\
            \phantom{1} + ScoreHMR w/o $\bb$     & \textbf{51.1} & 76.6\\
            \phantom{1} + ScoreHMR w/  $\bb$     & \textbf{51.1} & \textbf{76.5}\\
            \bottomrule
        \end{tabular}
    }
    \vspace{-2mm}
    \caption{ScoreHMR with and without the inclusion of SMPL shape parameters $\bb$. Numbers are PA-MPJPE in mm. Parenthesis denotes the number of body joints used to compute PA-MPJPE.}
    \label{tab:dm_betas}
    \vspace{-2mm}
\end{table}

\vspace{1mm}
\noindent\textbf{Refinement from HMR 2.0a.}
The Table below shows the PA-MPJPE of model fitting on 3DPW test set, starting from HMR 2.0a regression. Only ScoreHMR can quantitatively improve the performance of HMR 2.0a (by $4.5\%$).
\begin{table}[h]
    \centering
    \resizebox{0.95\columnwidth}{!}{
    \begin{tabular}{c|ccc}
        HMR 2.0a & +ScoreHMR & +ProHMR-fitting & +SMPLify \\ 
        \hline
        44.5 & \textbf{42.5} & 54.9 & 52.5 \\
    \end{tabular}
    }
\end{table}

\section{Additional Qualitative Results}
\phantom{11}
In Figure~\ref{fig:supp_fit} we include additional qualitative examples of model fitting, comparing our proposed approach with ProHMR-fitting~\cite{kolotouros2021probabilistic} and SMPLify~\cite{bogo2016keep}. Our approach achieves more faithful reconstructions than the baselines. We observe that in the case of missing keypoint detections (\eg, example with truncation in last row) SMPLify results in body orientation errors.

\begin{figure*}[t]
  \centering
  \includegraphics[width=0.9\linewidth]{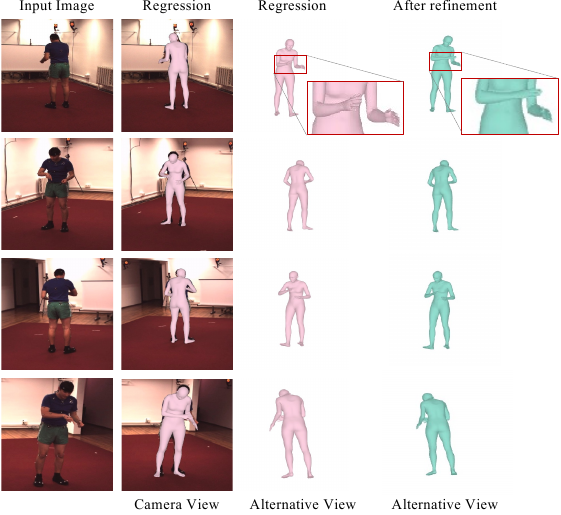}
  \caption{ \textbf{Multi-view refinement.} Refinement with multiple views fixes the 3D pose of the right hand, which is self-occluded in the first view (first row).
  }
  \label{fig:supp_multiview}
\end{figure*}

In Figure~\ref{fig:supp_multiview} we illustrate the effectiveness of our approach in consolidating information from multiple views in order to improve the 3D pose of a human. The initial view (first row of Figure~\ref{fig:supp_multiview}) presents challenges with occluded hands, resulting in inaccurate pose estimate for the hands. Multiple view fusion with our proposed approach results in a more faithful estimation of the true pose.

\begin{figure*}[t]
  \centering
    \includegraphics[width=\linewidth]{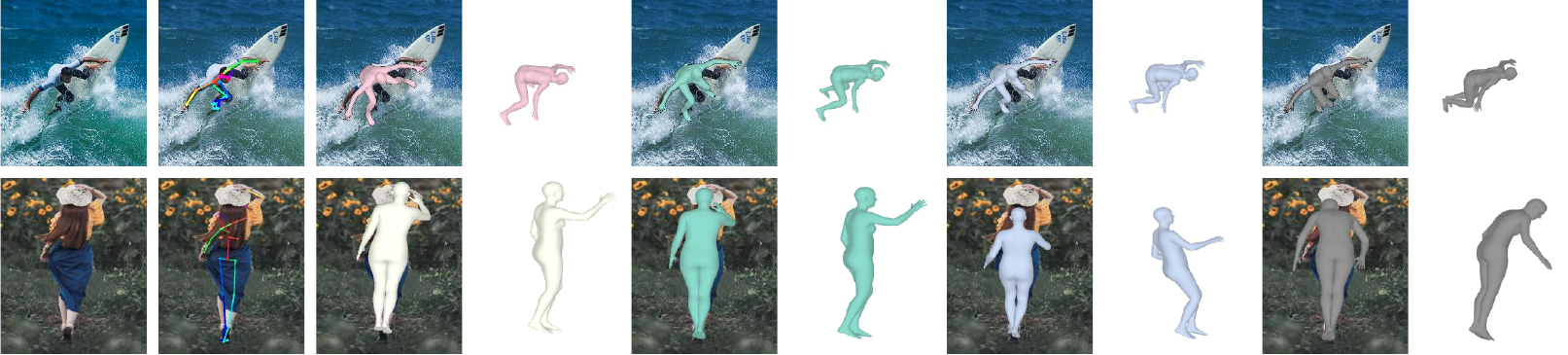}
  \caption{ \textbf{Failure cases of model fitting.} Pink: ProHMR regression. White: HMR 2.0b regression. Green: Regression + ScoreHMR (ours). Blue: Regression + ProHMR-fitting. Grey: Regression + SMPLify. While all methods encounter challenges when incorrect keypoints are detected, our image-conditioned diffusion model tries to keep the 3D pose aligned with the available image evidence.
  }
  \label{fig:supp_fit_fail}
  \vspace{4mm}
\end{figure*}

We present some failure cases of our method in Figure~\ref{fig:supp_fit_fail}. Our approach can fail when there are wrong keypoint detections. Optimization-based methods fail in that case too as we show in Figure~\ref{fig:supp_fit_fail}.

Finally, we demonstrate our approach on video sequences from the validation split of PoseTrack~\cite{andriluka2018posetrack} and others. We use predicted tracks from 4DHumans~\cite{goel2023humans}. We encourage viewing video results on the \href{https://statho.github.io/ScoreHMR}{project page}.

\end{document}